\documentclass[graybox, natbib, nosecnum]{svmult}
\bibpunct{(}{)}{;}{a}{}{,} 

\usepackage{mathptmx}       
\usepackage{helvet}         
\usepackage{courier}        
\usepackage{type1cm}        
\usepackage{natbib}

\usepackage{makeidx}         
\usepackage{graphicx}        
\usepackage{multicol}        
\usepackage[bottom]{footmisc}
\usepackage[normalem]{ulem}	
\usepackage{hyperref}  
\usepackage{soul}   

\begin{document}

\title*{Service-Oriented Robotic Computing for Cloud Robotics}
\author{Anis Koubaa}
\institute{Anis Koubaa \at Prince Sultan University, Saudi Arabia/CISTER Research Unit, Portugal/Gaitech Robotics, China, \email{akoubaa@psu.edu.sa}}
%
%
\maketitle
\section{Synonyms}
Service-Oriented Software Architecture for Cloud Robotics

\section{Definitions}

\begin{itemize}
	\item \textbf{Service-Oriented Architecture (SOA):} it is a software design methodology based on the interaction of software components with each other through service interfaces. It allows to build complex software systems that produce and consume services with each other.
	\item \textbf{Web services:}  it is an instantiation of service-oriented architecture that provides a standard approach to expose services through the Internet using Web technologies such as XML and JSON. It is based on the Simple Object Access Protocol (SOAP) protocol to exchange data, and on the Web Services Description Language (WSDL) for service description.
	\item \textbf{RESTful  Web services:}  REpresentational State Transfer (REST) is an architecture style of Stateless Web services based on the HTTP protocol to provide interoperability between machines on the Internet. A \textit{stateless} Web service means that each request is treated independently of its previous or future requests.  		
	\item \textbf{Cloud:} In computing, the cloud refers to a large data center with several powerful computer machines organized into clusters and connected through networks to provide computing and storage resources. 
	\item \textbf{Cloud Robotics:} it refers to the concept of integrating robots into cloud environments to take benefits of cloud resources in processing computation-extensive applications of robots on the cloud. 
	\item \textbf{Virtualization:} it consists in creating virtual environments of resources from physical resources to allow their sharing and virtual access. In the context of robotics computing, virtualization refers to the techniques that allow seamless access to robots anytime and anywhere through Web services and network interfaces. 
	\item \textbf{Computation offloading:} In the context of cloud robotics, computation offloading refers to migrating extensive computations from the robot to the cloud to leverage the high computing resources of the cloud. 
\end{itemize}

\section{Abstract}
In this article, we present an overview on the use of service-oriented architecture and Web services in developing robotics applications and software integrated with the Internet and the Cloud. This is a recent trend that emerged since 2010 from the concept of cloud robotics, which leverages the use of cloud infrastructures for robotics applications following a service-oriented architecture approach. In particular, we distinguish two main categories: (\textit{i.}) virtualization of robotics systems and (\textit{ii.}) computation offloading from robots to cloud-based services. We discuss the main approaches proposed in the literature to design robotics systems through the Web and their integration to the cloud through service-oriented computing framework.

\section*{Why Service-Oriented Robotic Computing}
Service-oriented robotic computing leverages the use of Service-Oriented Architecture (SOA) to build robotics software systems at large scale and through the Internet. It consists in using Web technologies to define services through which (\textit{i.}) robots can be accessed and/or, (\textit{ii.}) robots can access the resources of other machines that are typically deployed on the cloud. This paradigm allows to expand the use of robots for a larger number of users because it becomes easier to access them through user-friendly interfaces. It also promotes the interaction between robots and other machines through the Internet using the concept of \textit{service}. 
Indeed, the use of robots at very large public has been restricted so far due to several factors, including: 
\begin{enumerate}
	\item Robots have typically been used as standalone systems for very specific and dedicated missions in controlled environments, such as in industrial manufacturing, hospitals or homes. Even in the case of cooperative and multi-robots applications, robots communicate with users or robots to perform specific pre-programmed and pre-defined missions, but are isolated from the external world, and cannot learn from other contexts.
	\item The complexity of configuration and maintenance of the robots makes the use of robots challenging for non-technical and non-computer savvy users.
	\item The relatively high cost of sophisticated service robots in particular for professional use and some domestic applications.
\end{enumerate}

Addressing the aforementioned factors would help to promote the expansion of robots use at a much larger scale. Indeed, there is an increasing demand and interest in using the Internet infrastructure as a means to promote robots and robotics applications \textit{as services} in several perspectives: first, through exposing robotics resources as Internet services to end-users; second to foster the world-wide interaction between robots themselves and between robots and users through the Internet. In addition, with the emergence of cloud computing, service-oriented architecture and Internet-of-Things (IoT), robots may also take benefit from the huge computing resources available through the Internet to increase its processing capabilities for computation intensive applications. The cloud robotics paradigm, coined in 2010 by James Kuffner \cite{Kuffner2010}, seems to be the missing building block towards jumping to a new frontier of the public use of robots. 

Service-Oriented Architecture is a basic building block in the integration of computing systems that allow heterogeneous systems to expose their resources and use the resources of other systems through \textit{services}. A service can be seen as an abstract software interface between two end-systems that allow them to exchange messages. Recently, this concept has been applied widely between robotic systems and other systems/users interacting with them, which contributed to several approaches of service-oriented robotics systems. This chapter presents the main concepts for building service-oriented computing frameworks, and provides a review of the main approaches. 

\section*{Two Major Categories}
The integration of robotics applications through the IoT and the Cloud brings several benefits to robotics computing systems. 
We categorize service-oriented robotic computing from two perspectives: (\textit{i.}) \textbf{virtualization}: which means providing seamless access to robots through service interfaces (\textit{ii.}) \textbf{computation offloading} also know as \textbf{remote brains}, where computation is offloaded from robots to the cloud through service interfaces.

Robots typically need to process a large amount of data coming from its sensors to transform these data into a knowledge that allows the robot to execute specific actions, such as 3D map reconstruction from different sensor data sources, or object recognition and tracking using computer vision and 3D point cloud extensive computations. These types of applications typically require high processing capabilities, and consumes considerable energy. However, the onboard processing capabilities of robots, in particular low-cost platforms, might be insufficient to handle computation and storage intensive tasks. For this reason, outsourcing computations to more powerful and resource-abundant devices is a current trend nowadays.

In what follows, we provide an overview of the main contributions related to the two aforementioned categories, we outline the key challenges addressed in each paper, and we discuss the proposed approaches.

\section*{Virtualization}
The virtualization of access to robots means that users are able to access robots anywhere and anytime through simple-to-use interfaces, typically through the Web or mobile applications. Usually, these interfaces rely on services' abstraction layers that connect robots to users or other systems based on Web service technologies. 
\subsection*{Virtualization Technologies}
The main categories of Web services are \textit{SOAP web services} and \textit{REST Web services}. The reader may refer to \cite{Pautasso:2008} for a thorough discussion and analysis about the comparison between REST and SOAP Web services.

\begin{figure}[h]
	\centering
	\includegraphics[width=0.99\textwidth]{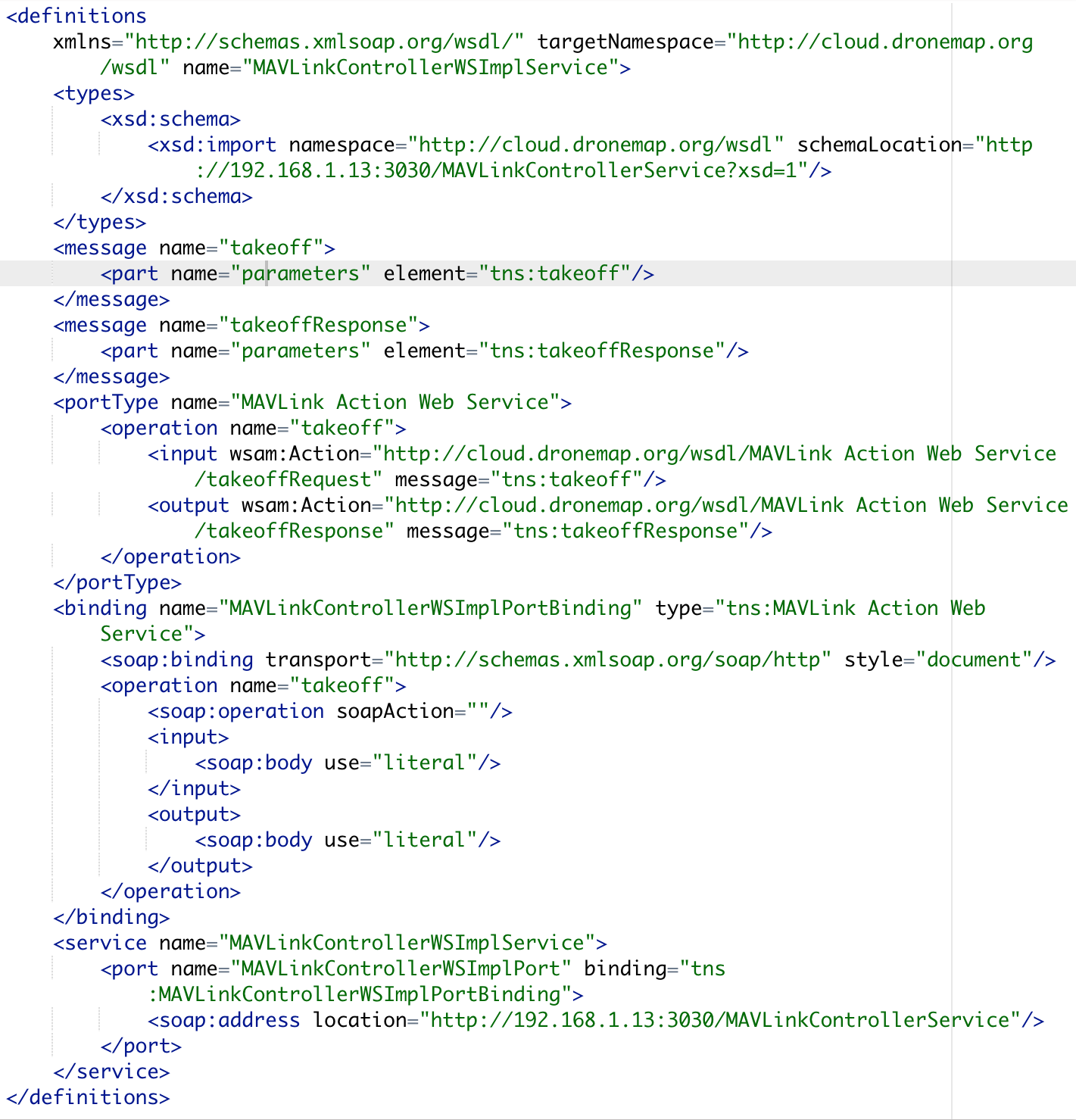}
	\caption{Example of a WSDL document of SOAP Web Service to control a drone }
	\label{wsdl}
\end{figure}

On the one hand, SOAP Web services provide a well-defined contract-based specification of the services. Services are described in an XML-based language called the Web Services Description Language (WSDL). Messages are exchanged between the client and the server through a standardized message exchange protocol using the SOAP envelope. To illustrate the concept, Figure \ref{wsdl} presents an excerpt of WSDL document of a SOAP Web service used to access and control a drone. We observe that the WSDL document defines the messages exchanged and their types, in addition to the different operations that can be executed as a service. For example, in the WSDL of Figure \ref{wsdl}, we identify a Web service called \texttt{MAVLink Action Web Service}, which exposes several operations as Web service methods that can be invoked by the Web service client applications. In this excerpt, we observe the \texttt{takeoff} Web service method (i.e. operation) which can be invoked to take-off the drone, etc. For every operation, a message is defined for the request and another is defined for the response, where a message may or may not have a parameter. For example, in Figure \ref{wsdl}, the \texttt{takeoff} operation defines the  \texttt{takeoff} message for the request, and the  \texttt{takeoffResponse} message for the response. Other operations are also defined in the same way. The WSDL document represents a contract-like specification that defines all operations available as services for the developers and end-users. Thus, the client application needs to implement client methods that invoke these operations on the server.  

\begin{figure}[h]
	\centering
	\includegraphics[width=0.99\textwidth]{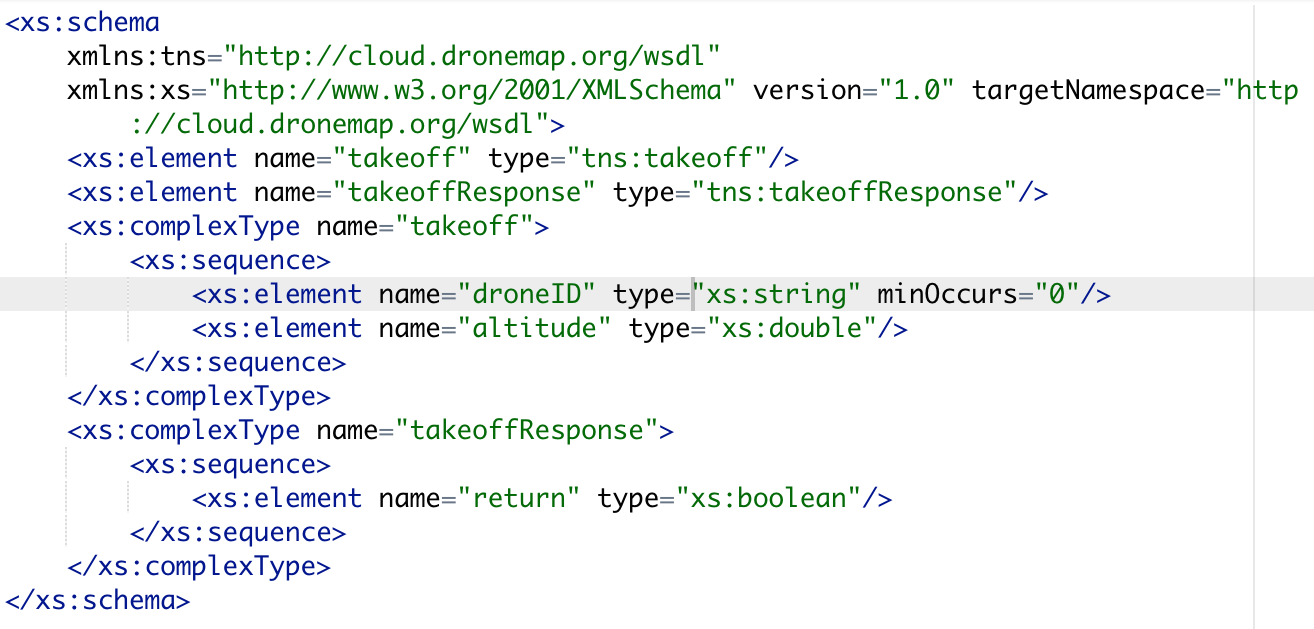}
	\caption{Example of a XSD document of SOAP Web Service to control a drone }
	\label{xsd}
\end{figure}

A message in SOAP can have parameters. The structure of a message and its parameters are defined in the XML Schema Definition (XSD) document, which describes the structure of the WSDL document in XML. Figure \ref{xsd} illustrates the XSD document for the WSDL document of Figure \ref{wsdl}. For example, we observe that the \texttt{takeoff} message uses two parameters, which is the ID of a drone to takeoff, and also the desired altitude. The \texttt{takeoffResponse} message returns a boolean response to notify about the success of the operation. According to the programming language API, the application developer will consider the WSDL document as a contract to develop the client methods that will invoke the Web methods provided by the SOAP Web service to execute the actions of interest. It has to be noted that SOAP Web services are  programming-language independent and platform-independent. 

On the other hand, REST is a more lightweight Web service solution that does not define a formal service specification as with SOAP. The interaction between the client and the server is ensured through the basic HTTP protocol. The programming abstractions of REST Web services are different from those of SOAP. While SOAP is based on a concrete description of Web services in the WSDL document, REST refers to services through their Uniform Resource Identifier (URI), which are simply Web links defining paths to resources located in the Web server. Every operation in REST is a resource that has a dedicated URI to it in the following format. 
\begin{verbatim}
scheme:[//[user[:password]@]host[:port]][/path]
\end{verbatim}

Considering the example above of the \texttt{takeoff} Web service method expressed as REST, the access to this operation can be defined by a URI in the form:

\begin{verbatim}
http://www.domain.com/drone/takeoff/id/1/altitude/20/
\end{verbatim} 

This URI refers to the Web service operation \texttt{takeoff} in the public server \texttt{www.domain.com}, with a resource path \texttt{/drone/takeoff/} to takeoff the drone of id 1 to an altitude 20 meters. It is clear that RESTful Web service are simpler to use and to define as compared to the contract-based approach of SOAP Web services. 

\begin{figure}[h]
	\centering
	\includegraphics[width=0.99\textwidth]{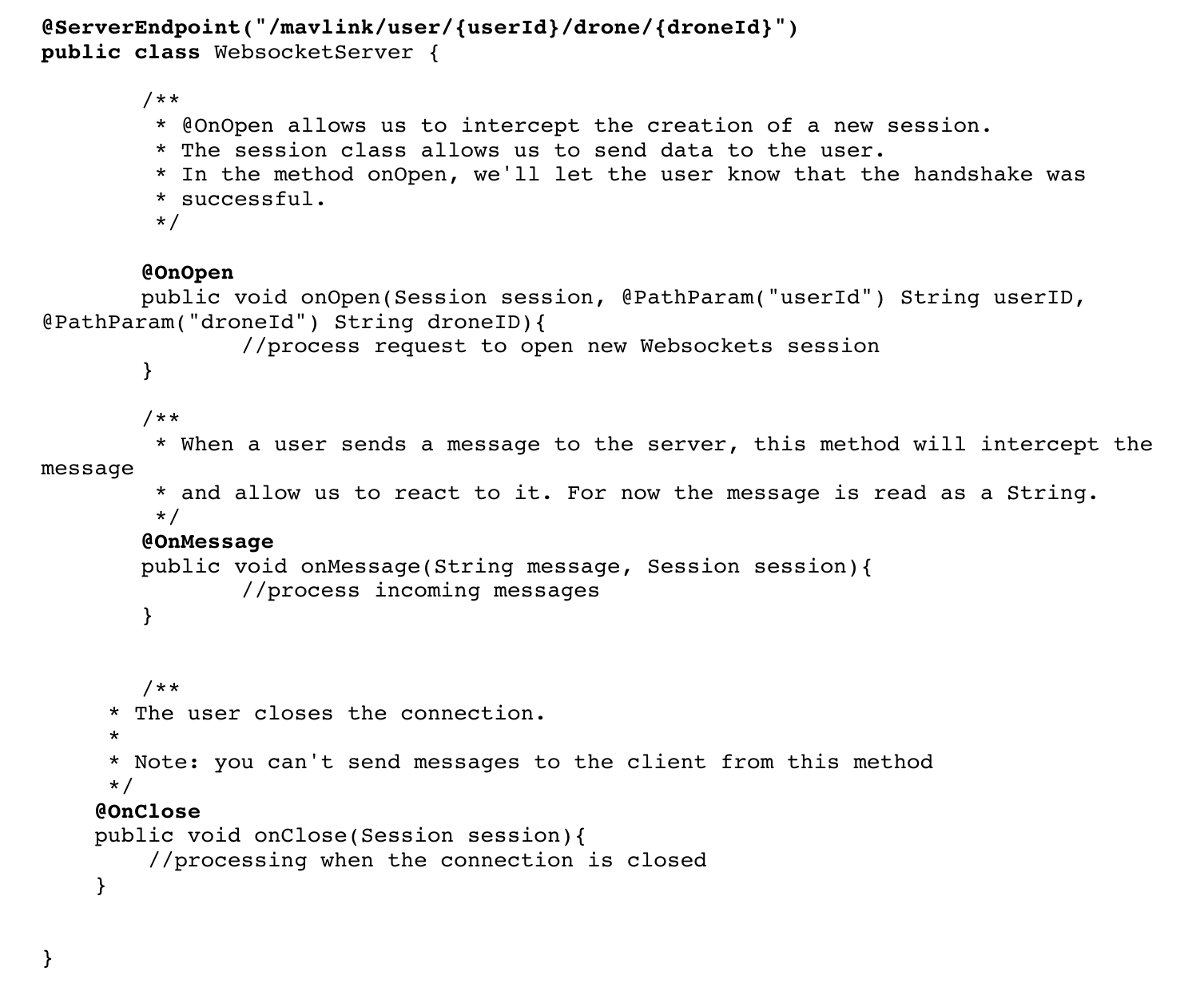}
	\caption{Example of a Websockets class in Java}
	\label{websockets}
\end{figure}

Websockets represent another technology that is widely used in the Web for a standard and platform-independent message exchange between a client and a server. It is a core technology for the IoT and real-time streaming applications. It is a bi-directional communication protocol between a client and a server, such that both parties can exchange messages simultaneously in full-duplex mode. A connection is open between the client and the server following a handshake process, then both the server and the client can exchange messages asynchronously and reliably in real-time until the connection is explicitly closed by any of the communicating entities.  The advantage of Websockets is that it is supported by major programming languages such as Java, C/C++, Python, Ruby, JavaScript, PHP, to name a few. This allows for an easy integration of heterogeneous robotics systems. Websockets were used by Osentoski et al. and Crick et al. in \cite{Osentoski:2011,Crick11rosbridge} respectively to connect ROS-enabled robots to the Internet. It was also used by Koubaa et al. in \cite{dronemap, dp2019, dronetrack2018} for real-time streaming of MAVLink messages \cite{mavlink} between drones and client applications, in addition to SOAP and REST Web services. 
Figure \ref{websockets} represents an example of a Websockets class template in Java. The structure of the code is similar in other programming languages. It can be observed that Websockets are defined like Web services by an endpoint entry, which defines the path to the Websockets resource, similarly to a REST URI. In this example, the resource is called \texttt{/mavlink/user/{userId}/drone/{droneId}}. The event \texttt{@onOpen} processes incoming connections initialization requests, and creates and manages sessions. The event \texttt{@onMessage} intercepts incoming messages and processes them according to the application logic. The event \texttt{@onClose} manages connection closing events. For example, in \cite{dronemap, dp2019}, the Web client application uses a Websockets connection with the cloud to receive the stream of data representing states of a drone through the cloud in real-time, and displays it using JavaScript on a Web browser.

\subsection*{Recent Contributions on Robotics Virtualization}
There are several works that aim at virtualizing access to robots using Web services and SOA. Many of these works target the Robot Operating System (ROS), as a main development framework for robotics applications. In \cite{Osentoski:2011,Crick11rosbridge}, the authors propose the \texttt{rosjs}, and the \texttt{rosbridge} middleware. These works represent a milestone in the integration of ROS into the Web and the Internet. The motivation behind \texttt{rosbridge} and \texttt{rosjs} is mainly two fold: (1) to use commonly available Internet browsers for non-roboticians users to interact with a ROS- enabled robot (2) to provide Web developers with no background in robotics with simple interfaces to develop client applications to control and manipulate ROS-enabled robots.

In \cite{Koubaa2014}, the author proposes, RoboWeb, a robot virtualization service-oriented architecture based on SOAP Web services. The objective of RoboWeb is to develop a remote robotics lab that can seamlessly be accessed by researchers and students anywhere and anytime. RoboWeb allows to monitor and control robots through Web services. The service-oriented architecture is composed of three main layers: (1) the web interface layer, which uses rosPHP API to access ROS-enabled robots through the web, (2) the service broker that defines a middleware that enables interaction between users and robots, and (3) the robot itself, which must support ROS. A prototype was implemented and tested on the Turtlebot robot. The limitation of this work is that SOAP Web services were not integrated into ROS ecosystem but were developed externally to the ROS ecosystem. Also, no REST Web service interface was proposed. 

In \cite{koubaa15}, the author addresses the problems of \cite{Koubaa2014} and presents ROS Web services. The objective was to expose ROS resources as Web services. The innovation of this paper was to design an object-oriented software architecture to integrate Web Services into ROS and exposes its resources (i.e. ROS topics and ROS services) as Web services. Both SOAP and REST Web services were proposed. A prototype implementation was used to demonstrate how ROS Web services promote portability, reusability, and interoperability of ROS-enabled robots with client applications. 

\begin{figure}[h]
	\centering
	\includegraphics[width=0.95\textwidth]{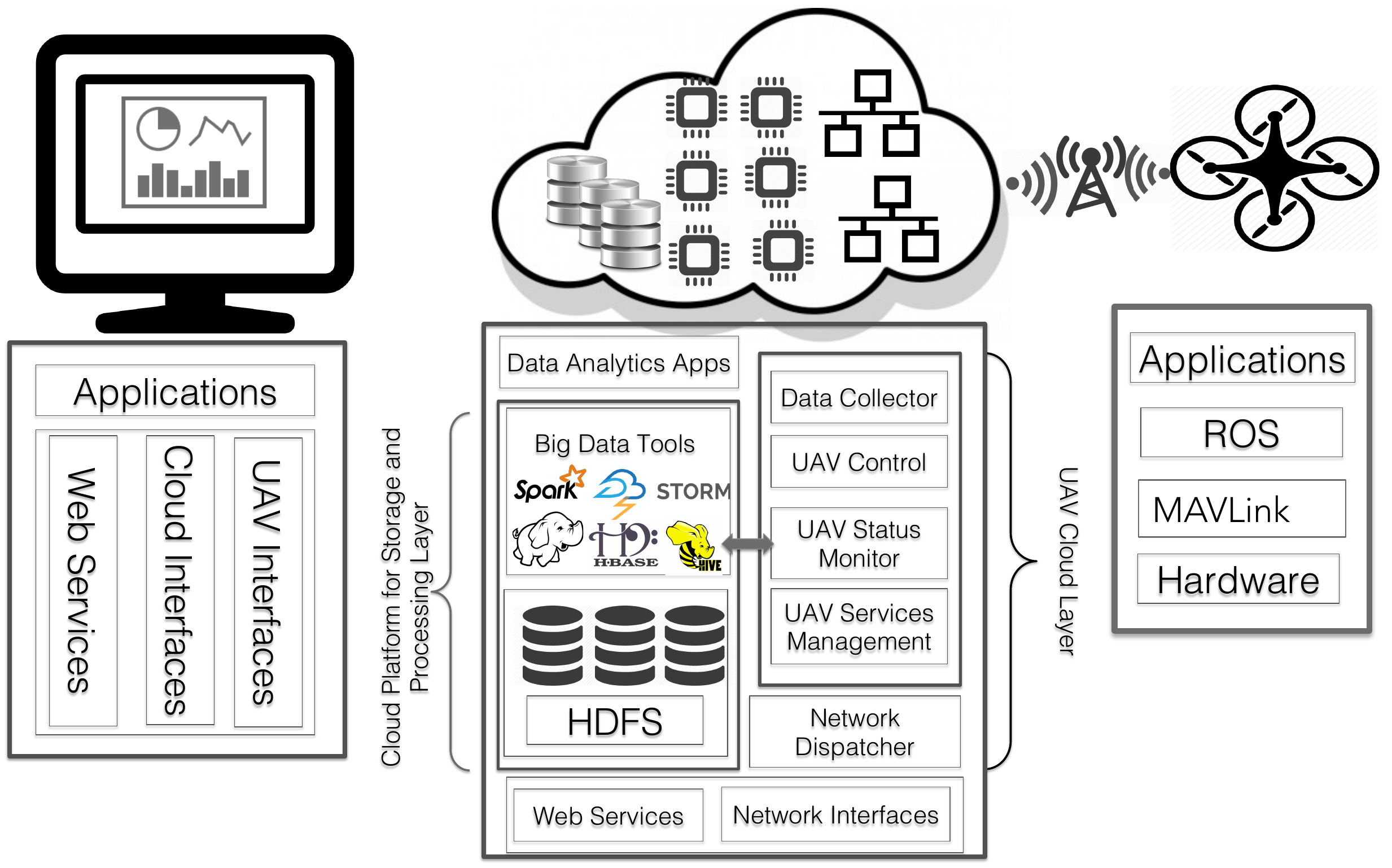}
	\caption{DroneMap System Architecture: Abstraction Layers \cite{dronemap}}
	\label{dp}
\end{figure}

In \cite{dronemap,dp2019}, Dronemap Planner (DP) a service-oriented cloud-based drones management system was proposed for MAVLink-based drones. The paper also introduces the concept of Internet-of-Drones (IoD) and discussed their functional and non-functional requirements. In \cite{dronetrack2018}, the authors evaluated the real-time performance the Internet of Drones using a GPS-based tracking application.
Also in \cite{gharibi2016internet}, the authors proposed their vision and architecture for IoD. 
DP provides seamless access to drones through SOAP and REST Web service technologies, schedules their missions, and promotes collaboration between them. The architecture is illustrated in Figure \ref{dp}. The Dronemap Planner system is composed of three abstraction layers. The first layer is the drone, which supports ROS and the MAVLink protocol to communicate through the Internet with the cloud. The second layer is the cloud manager which links the drones to the end-users applications. The third layer is the end-user applications, which is used to monitor and control the drone through the Dronemap Planner cloud system. 

The main contribution of \cite{dronemap,dp2019} consists in deploying a cloud platform that contains a proxy server that relays between drones and users. In addition, REST and SOAP Web services interfaces are used to allow the user to interact with the cloud to send commands to the drones. Experimental deployment demonstrates that Dronemap Planner is effective in virtualizing the access to drones over the Internet, and provides developers with appropriate APIs to easily program drones' applications. 

\begin{figure}[ht]
	\centering
	\includegraphics[width=0.95\textwidth]{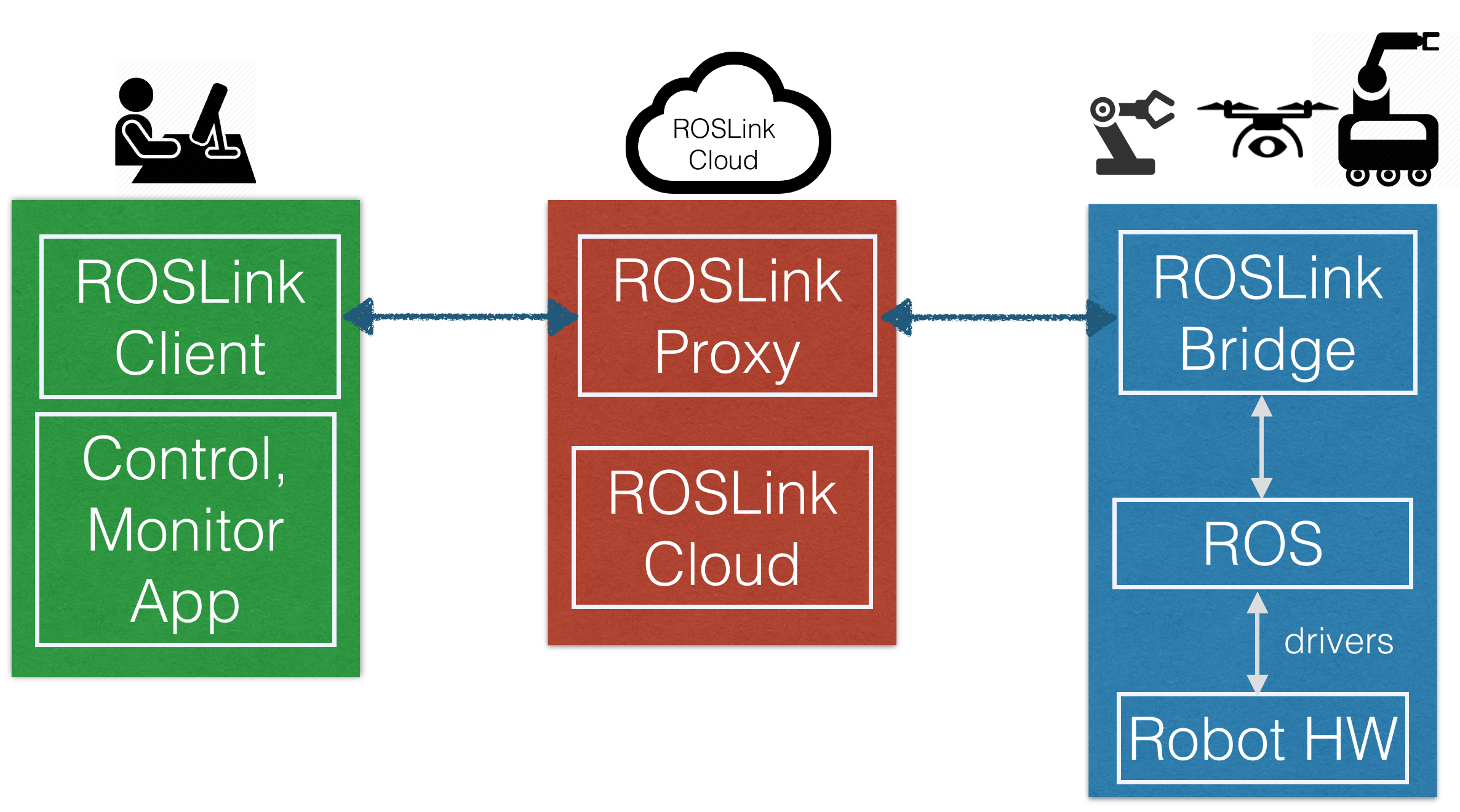}
	\caption{ROSLink Architecture \cite{roslink2017}}
	\label{roslink}
\end{figure}

In \cite{roslink2017}, the authors propose \texttt{ROSLink} as a new alternative to integrate ROS into the Internet-of-Things. The authors started from the observation that previous works in the literature proposed to develop \textit{robot-centric} approach meaning that a Web server is developed in the ROS robot machine to deliver data from the robot to the clients. This was demonstrated to restrict the scalability of the system as the server is centralized in the robot itself. In addition, the deployment of robot-centered solutions on the Internet is rather difficult as the robot needs to have a public IP address or be accessible through a NAT forwarding port when it is inside a local area network.
The idea behind ROSLink is three-tier client/server model, where the clients are implemented in both the robot and the user, whereas the server is deployed on a cloud infrastructure on the Internet with a public IP address.  The cloud manager of the Dronemap Planner system \cite{dp2019} was extended to also support the ROSLink protocol. The \texttt{ROSLink} architecture is presented in Figure \ref{roslink}. The three-tier ROSLink architecture represents a possible approach of implementing service-oriented computing system for robotics that allows for seamless access of ROS-enabled robots through the Internet. 

The authors describe the communication protocol specification of \texttt{ROSLink} and evaluate its performance using qualitative and quantitative analysis. The communication protocol is based on JSON serialized messages exchanged between the robots, the cloud, and the users. The performance study was conducted on an open-loop spiral trajectory control application using Turtlesim simulator for both ROS and ROSLink deployed on the cloud. 

The strength of the ROSLink service-oriented architecture is that it easily promotes the integration of ROS-enabled robots into the Internet-of-Things using Web services. However, the performance of robotics applications using ROSLink heavily depends on the network quality-of-service. This remains a challenge that must be thoroughly investigated.  

In \cite{Mahmoud:2014:ICUAS}, the authors present a service-oriented architecture for collaborative drones. They propose a mapping model between cloud computing resources and drone resources. In addition, two types of services including essential services and customized services were proposed. The paper only provides a high-level description of the system architecture, components and services without any specific details on their implementation. 

In \cite{Mahmoud:2015:CTS}, the authors of \cite{Mahmoud:2014:ICUAS} extended their approach  and designed a REST architecture using a Resource-Oriented Architecture (ROA) model to represent the services and resources of drones. In addition, a broker that dispatches mission requests to available UAVs was proposed. The broker is responsible for managing the UAVs, their missions and their interactions with the client. The authors validated their proposal on a simple Arduino board that emulates a UAV and its resources, which represents a limitation, because it does not show a comprehensive proof of concept on real drones.

Some other recent approaches consider more generic models for SOA to access robots without the use of Web services.  In \cite{Brugali:2014:SOSE}, Brugali et al. propose the Task Component Architecture (TCA) for the seamless integration of the Service Component Architecture (SCA) into robotic software control systems to execute asynchronous tasks. The authors also integrate TCA with ROS using a \texttt{ROSProxyNode} developed with \texttt{ROSJAVA}. 

On the other hand, large cloud computing service provided developed their own platforms for developing robotics applications through the cloud. As a matter of fact, Amazon Web Service (AWS) provides the AWS RoboMaker service \cite{RoboMaker2019} that enables users to develop and deploy mobile robots applications through AWS cloud services. It supports ROS-enabled robots and also provides computation offloading services to perform machine learning and data analytics for data collected from the robots.

\section*{Computation Offloading}

The idea of computation offloading came from the fact that robots, in particular with low-cost systems, have limited energy, processing and storage capabilities. In fact, robots can process sensor data using multiple onboard processors. Nonetheless, some other types of robots namely drones or even ground mobile robots may have limited onboard computing and storage capabilities, which is typically due to the constraints on their dimensional, power and payload requirements. On the other hand, robotic applications typically require massive computations and storage requirements, in particular applications using computer vision, signal processing, real-time map building, location, navigation, to name a few. Thus, instead of overwhelming robots with such intensive computation, it is transferred to a remote server located on a cloud platform for processing and then returning the results or actions to the robot. This is the concept of computation offloading. 

\par In the recent years, there have been several attempts to integrate robots with the cloud through service-oriented and web interfaces. 
It has to be noted that this becomes possible nowadays with the great expansion of available bandwidth through the Internet allowing the real-time exchange of data with high data rate requirements through high-speed networks either through landlines or optical fiber cables or even wireless communication, which now supports very high bandwidths up to 200 \texttt{Mbps} and more.  Furthermore, the evolution of cloud computing platforms has made it possible to leverage abundant computing resources on the cloud for processing intensive applications.

\par One of the first contributions in this area is the DaVinci project reported in \cite{4rb}. The DaVinci server acts as a proxy that relays the access between robots and users. It also supports the Hadoop distributed File System (HDFS) and Robot Operating System (ROS). The objective of DaVinci is to offload intensive computation from the resources of the robots to a back-end cluster system in the cloud. The idea was to investigate the possibility of parallel execution of complex robotic algorithms, and to apply it to the FastSLAM algorithm as a proof of concept. The deployment did not consider network latencies and delays, which limits the results to ideal operational conditions.

\par 
In \cite{2rb}, Hunziker et al. proposed Rapyuta in the context of the ROBOEARTH project. In this project, a cloud engine was devised to promote collaboration between robots, managing robotic resources and share knowledge among robots through an open source cloud robotics framework. Rapyuta is deployed on Amazon Web services cloud where computation is offloaded from the robots for processing on the cloud. The cloud platform provides a secure and elastic computing environment, and in addition compatibility with ROS. Communication between robots uses the Websockets protocol to provide a full-duplex communication channel between robots. 

In \cite{OpenEASE}, the authors proposed OpenEASE, which is a knowledge-based framework that allows to shares data between robots and humans. The system is composed of a database that contains a large set of semantically annotated data collected from humans and robots during complex manipulation tasks. It contains all information about the manipulation tasks such as the environment of the mission, the manipulated objects, the tasks executed and the behavior generated. A Web browser and Websocket API interfaces are used to send queries and vizualize the manipulation mission.

In \cite{moh2014}, the authors addressed the problem of collaborative 3D mapping on the cloud using low-cost robots. They proposed an architecture based on the Rapyuta cloud engine to process visual odometry coming from robots, which will perform parallel optimization and merging of maps produced by other robots. The results of optimization are pushed back to the robots. The authors evaluated the performance of cloud-based collaborative 3D mapping in terms of accuracy,  bandwidth and storage requirements, and execution time.
The paper demonstrated that the cloud-based architecture contributes to increasing the number of cooperative low-cost robots performing accurate mapping and localization. It has been shown that the experimental implementation resulted in maps with quality comparable to those performed by more expensive robot hardware. The bandwidth requirement is as small as 0.5 MB/sec, which is within the range of typical wireless networks. This prototype demonstrates how the integration of robots with cloud computing opens a new horizon for low-cost robots to perform more complex computation-intensive tasks. 

In \cite{RCoff6} the authors proposed a Cloud Enabled Robotics System (CERS) where robots outsource their heavy computations to a server deployed in a cloud platform. The paper also discussed security aspects while integrating robots with the cloud. The proposed system was evaluated with a real-time video tracking application and compared the performance obtained by virtual machines and physical machines clusters. 


In \cite{Bekris} the authors investigated the opportunities in industrial automation with the advanced in cloud computing. The paper proposes to use the cloud to solve complex problems related to motion planning of manipulators. The computation load is split between the robot local machine and the cloud. The evaluation study demonstrated the effectiveness of using the cloud to compute roadmap data structures. 

Reference \cite{Salm2015} conducted a study to demonstrate the effectiveness of computation offloading to the cloud in the context of vision-based navigation assistance of a service robot. Data is offloaded from a stereo camera sensor on the robot to a private cloud built using OpenStack. The experimental evaluation demonstrated that computation offloading improved the navigation experience of the robot as compared to all processing being done onboard. 

In \cite{lei2016}, Lei et al. propose the Cloud Robotics Visual Platform(CRVP) for offloading computer vision application from the robot to the cloud. The Hadoop Map/Reduce framework was used to reduce the time of learning and recognition processes. The authors designed a service-oriented architecture to build the recognition engine and deployed on Amazon Elastic Compute System (ECS) on Amazon Web services cloud. The experiments were conducted on a face recognition application applied to 600 images, with a recognition execution time around 60 ms.


\section*{Recommended Readings}
In the previous section, we presented an overview of the main related works in the literature for each category of service-oriented robotic computing systems. However, this list is not exhaustive and there far more works in the literature. In this section, we provide pointers to some recommended readings for the reader to get more insights into the cloud robotics area. 

The survey in \cite{Chaari2016} presents a comprehensive overview of cyber-physical clouds including robots. This survey is a good starting point to take a bird-eye-view on efforts being done in the integration of cyber-physical systems (namely robots, sensors and vehicles) into the Internet-of-Things and cloud. In addition, \cite{kehoe2015} provides a detailed survey on cloud robotics research and is organized into four categories, including Big Data, Cloud Computing, Collective Learning and Human Computation. 


\bibliographystyle{IEEEtran}
\bibliography{bib}

\begin{thebibliography}{10}
\providecommand{\url}[1]{#1}
\csname url@samestyle\endcsname
\providecommand{\newblock}{\relax}
\providecommand{\bibinfo}[2]{#2}
\providecommand{\BIBentrySTDinterwordspacing}{\spaceskip=0pt\relax}
\providecommand{\BIBentryALTinterwordstretchfactor}{4}
\providecommand{\BIBentryALTinterwordspacing}{\spaceskip=\fontdimen2\font plus
\BIBentryALTinterwordstretchfactor\fontdimen3\font minus
  \fontdimen4\font\relax}
\providecommand{\BIBforeignlanguage}[2]{{%
\expandafter\ifx\csname l@#1\endcsname\relax
\typeout{** WARNING: IEEEtran.bst: No hyphenation pattern has been}%
\typeout{** loaded for the language `#1'. Using the pattern for}%
\typeout{** the default language instead.}%
\else
\language=\csname l@#1\endcsname
\fi
#2}}
\providecommand{\BIBdecl}{\relax}
\BIBdecl

\bibitem{Kuffner2010}
J.~Kuffner, ``Cloud-enabled robots,'' in \emph{IEEE-RAS International
  Conference on Humanoid Robots}.\hskip 1em plus 0.5em minus 0.4em\relax IEEE,
  2010.

\bibitem{Pautasso:2008}
\BIBentryALTinterwordspacing
C.~Pautasso, O.~Zimmermann, and F.~Leymann, ``Restful web services vs. "big"'
  web services: Making the right architectural decision,'' in \emph{Proceedings
  of the 17th International Conference on World Wide Web}, ser. WWW '08.\hskip
  1em plus 0.5em minus 0.4em\relax New York, NY, USA: ACM, 2008, pp. 805--814.
  [Online]. Available: \url{http://doi.acm.org/10.1145/1367497.1367606}
\BIBentrySTDinterwordspacing

\bibitem{Osentoski:2011}
S.~Osentoski, G.~Jay, C.~Crick, B.~Pitzer, C.~DuHadway, and O.~C. Jenkins,
  ``Robots as web services: Reproducible experimentation and application
  development using rosjs,'' in \emph{Robotics and Automation (ICRA), 2011 IEEE
  International Conference on}, 2011.

\bibitem{Crick11rosbridge}
C.~Crick, G.~T. Jay, S.~Osentoski, B.~Pitzer, and O.~C. Jenkins, ``rosbridge:
  Ros for non-ros users,'' in \emph{International Symposium on Robotics
  Research (ISRR 2011)}, Flagstaff, AZ, USA, August 2011.

\bibitem{dronemap}
A.~Koubaa, B.~Qureshi, M.-F. Sriti, Y.~Javed, and E.~Tovar, ``{Dronemap
  Planner: A Service-Oriented Cloud-Based Management System for the
  Internet-of-Drones},'' in \emph{The 17th International Conference on
  Autonomous Robot Systems and Competitions (ICARSC 2017)}, April 2017.

\bibitem{dp2019}
A.~Koubaa, B.~Qureshi, M.-F. Sriti, A.~Allouch, Y.~Javed, M.~Alajlan,
  O.~Cheikhrouhou, M.~Khalgui, and E.~Tovar, ``{Dronemap Planner: A
  Service-Oriented Cloud-Based Management System for the Internet-of-Drones},''
  \emph{Ad Hoc Networks}, vol.~86, 2019.

\bibitem{dronetrack2018}
A.~Koubaa and B.~Qureshi, ``Dronetrack: Cloud-based real-time object tracking
  using unmanned aerial vehicles,'' \emph{IEEE Access}, vol.~PP, pp. 1--1, 03
  2018.

\bibitem{mavlink}
MAVLINK, ``{The MAVLINK Protocol}, website:
  http://qgroundcontrol.org/mavlink/start,''
  http://qgroundcontrol.org/mavlink/start, 2019.

\bibitem{Koubaa2014}
\BIBentryALTinterwordspacing
A.~Koubaa, \emph{Architecture of Computing Systems -- ARCS 2014: 27th
  International Conference, L{\"u}beck, Germany, February 25-28, 2014.
  Proceedings}.\hskip 1em plus 0.5em minus 0.4em\relax Cham: Springer
  International Publishing, 2014, ch. A Service-Oriented Architecture for
  Virtualizing Robots in Robot-as-a-Service Clouds, pp. 196--208. [Online].
  Available: \url{http://dx.doi.org/10.1007/978-3-319-04891-8_17}
\BIBentrySTDinterwordspacing

\bibitem{koubaa15}
------, ``{ROS As a Service: Web Services for Robot Operating System},''
  \emph{Journal of Software Engineering for Robotics}, vol.~6, no.~1, 2015.

\bibitem{gharibi2016internet}
M.~Gharibi, R.~Boutaba, and S.~L. Waslander, ``Internet of drones,'' \emph{IEEE
  Access}, vol.~4, pp. 1148--1162, 2016.

\bibitem{roslink2017}
A.~Koubaa, M.~Alajlan, and B.~Qureshi, ``{ROSLink: Bridging ROS with the
  Internet-of-Things for Cloud Robotics},'' in \emph{Springer Book of Robot
  Operating System (ROS), Volume 2}, May 2017.

\bibitem{Mahmoud:2014:ICUAS}
S.~Mahmoud and N.~Mohamed, ``{Collaborative UAVs Cloud},'' in \emph{Unmanned
  Aircraft Systems (ICUAS), 2014 International Conference on}, May 2014, pp.
  365--373.

\bibitem{Mahmoud:2015:CTS}
------, ``{Broker architecture for collaborative UAVs cloud computing},'' in
  \emph{Collaboration Technologies and Systems (CTS), 2015 International
  Conference on}, June 2015, pp. 212--219.

\bibitem{Brugali:2014:SOSE}
D.~Brugali, A.~Da~Fonseca, A.~Luzzana, and Y.~Maccarana, ``{Developing Service
  Oriented Robot Control System},'' in \emph{Service Oriented System
  Engineering (SOSE), 2014 IEEE 8th International Symposium on}, April 2014,
  pp. 237--242.

\bibitem{RoboMaker2019}
\BIBentryALTinterwordspacing
(2019) Aws {R}obomaker: Amazon cloud robotics platform. [Online]. Available:
  \url{https://aws.amazon.com/robomaker/}
\BIBentrySTDinterwordspacing

\bibitem{4rb}
R.~Arumugam, V.~R. Enti, L.~Bingbing, W.~Xiaojun, K.~Baskaran, F.~F. Kong,
  A.~S. Kumar, K.~D. Meng, and G.~W. Kit, ``Davinci: A cloud computing
  framework for service robots,'' in \emph{Robotics and Automation (ICRA), 2010
  IEEE International Conference on}, May 2010, pp. 3084--3089.

\bibitem{2rb}
G.~Mohanarajah, D.~Hunziker, R.~D'Andrea, and M.~Waibel, ``Rapyuta: A cloud
  robotics platform,'' \emph{IEEE Transactions on Automation Science and
  Engineering}, vol.~12, no.~2, pp. 481--493, April 2015.

\bibitem{OpenEASE}
\BIBentryALTinterwordspacing
G.~Bartels, M.~Beetz, D.~Bessler, M.~Tenorth, and J.~Winkler, ``How to use
  openease: An online knowledge processing system for robots and robotics
  researchers (demonstration),'' in \emph{Proceedings of the 2015 International
  Conference on Autonomous Agents and Multiagent Systems}, ser. AAMAS
  '15.\hskip 1em plus 0.5em minus 0.4em\relax Richland, SC: International
  Foundation for Autonomous Agents and Multiagent Systems, 2015, pp.
  1925--1926. [Online]. Available:
  \url{http://dl.acm.org/citation.cfm?id=2772879.2773507}
\BIBentrySTDinterwordspacing

\bibitem{moh2014}
G.~Mohanarajah, V.~Usenko, M.~Singh, M.~Waibel, and R.~{D'Andrea},
  ``Cloud-based collaborative {3D} mapping in real-time with low-cost robots,''
  \emph{IEEE Transactions on Automation Science and Engineering}, March 2014.

\bibitem{RCoff6}
L.~Bingwei, C.~Yu, B.~Erik, P.~Khanh, S.~Dan, and C.~Genshe, ``A holistic
  cloud-enabled robotics system for real-time video tracking application,'' in
  \emph{Future Information Technology}, ser. Lecture Notes in Electrical
  Engineering, vol. 276.\hskip 1em plus 0.5em minus 0.4em\relax Springer Berlin
  Heidelberg, 2014, pp. 455 -- 468.

\bibitem{Bekris}
K.~Bekris, R.~Shome, A.~Krontiris, and A.~Dobson, ``Cloud automation:
  Precomputing roadmaps for flexible manipulation,'' \emph{IEEE Robotics
  Automation Magazine}, vol.~22, no.~2, pp. 41--50, June 2015.

\bibitem{Salm2015}
J.~Salmeron-Garcia, P.~Inigo-Blasco, F.~D. del Rıo, and D.~Cagigas-Muniz, ``A
  tradeoff analysis of a cloud-based robot navigation assistant using stereo
  image processing,'' \emph{IEEE Transactions on Automation Science and
  Engineering}, vol.~12, no.~2, pp. 444--454, April 2015.

\bibitem{lei2016}
Y.~Lei, Z.~Fengyu, W.~Yugang, Y.~Xianfeng, Z.~Yang, and C.~Zhumin, ``Design of
  a cloud robotics visual platform,'' in \emph{2016 Sixth International
  Conference on Instrumentation Measurement, Computer, Communication and
  Control (IMCCC)}, July 2016, pp. 1039--1043.

\bibitem{Chaari2016}
\BIBentryALTinterwordspacing
R.~Chaari, F.~Ellouze, A.~Koubaa, B.~Qureshi, N.~Pereira, H.~Youssef, and
  E.~Tovar, ``Cyber-physical systems clouds: A survey,'' \emph{Computer
  Networks}, vol. 108, pp. 260 -- 278, 2016. [Online]. Available:
  \url{http://www.sciencedirect.com/science/article/pii/S1389128616302699}
\BIBentrySTDinterwordspacing

\bibitem{kehoe2015}
B.~Kehoe, S.~Patil, P.~Abbeel, and K.~Goldberg, ``A survey of research on cloud
  robotics and automation,'' \emph{IEEE Transactions on Automation Science and
  Engineering}, vol.~12, no.~2, pp. 398--409, April 2015.

\end{thebibliography}


\end{document}